%% file: main.tex
\documentclass[10pt,twocolumn,letterpaper]{article}

\usepackage{cvpr}              

\input{preamble}

\definecolor{cvprblue}{rgb}{0.21,0.49,0.74}
\usepackage[pagebackref,breaklinks,colorlinks,citecolor=cvprblue]{hyperref}

\usepackage{multirow}
\usepackage{tabularx}

\def\paperID{*****} 
\def\confName{CVPR}
\def\confYear{2024}

\DeclareMathOperator*{\argmax}{argmax}

\title{Investigating the Semantic Robustness of CLIP-based Zero-Shot Anomaly Segmentation}

\author{Kevin Stangl\thanks{Work done while at Intel Labs, USA. \\ 
This publication includes work that was funded by the Government under the Defense Advanced Research Projects Agency (DARPA) Guaranteeing AI Robustness against Deception (GARD) Program, Agreement \#HR00112030001.}\\
TTIC, USA\\
{\tt\small kevin@ttic.edu}
\and
Marius Arvinte \quad Weilin Xu \quad Cory Cornelius\\
Intel Labs, USA\\
{\tt\small firstname.lastname@intel.com}
}

\begin{document}
\maketitle

\begin{abstract}
Zero-shot anomaly segmentation using pre-trained foundation models is a promising approach that enables effective algorithms without expensive, domain-specific training or fine-tuning. 
Ensuring that these methods work across various environmental conditions and are robust to distribution shifts is an open problem. 
We investigate the performance of WinCLIP \citep{jeong2023winclip} zero-shot anomaly segmentation algorithm by perturbing test data using three semantic transformations: bounded angular rotations, bounded saturation shifts, and hue shifts. 
We empirically measure a lower performance bound by aggregating across per-sample worst-case perturbations and find that average performance drops by up to $20\%$ in area under the ROC curve and $40\%$ in area under the per-region overlap curve. 
We find that performance is consistently lowered on three CLIP backbones, regardless of model architecture or learning objective, demonstrating a need for careful performance evaluation.
\end{abstract}

\maketitle

\section{Introduction}
Visual anomaly segmentation is a challenging and important task, especially in manufacturing line applications, where objects have to be inspected in a fast, automated way, that is robust under a varied number of environmental conditions \citep{bergmann2019mvtec,zou2022spotthedifference}. 
Anomalies manifest as fabrication defects (scratches, chips) and localizing them using fine-grained segmentation allows for interpretability of different types of defects \citep{bergmann2019mvtec}.
Given the variety of items that are inspected, collecting large, annotated training sets for each object is prohibitive. 
Even if data would be available, training and deploying separate models for each object is costly.

\begin{figure}[t]
    \centering
    \includegraphics[width=1\linewidth]{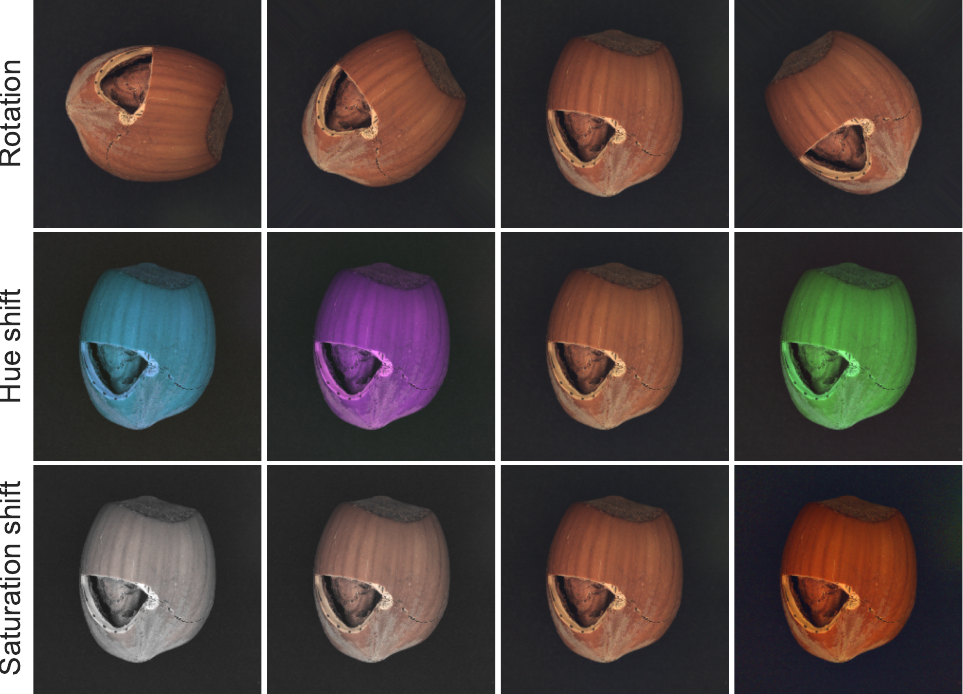}
    \caption{Effects of the three augmentations applied to the same anomalous (a large crack in the shell of a hazelnut) MVTec sample. The third column represents the original sample.}
    \label{fig:main_aug_examples}
\end{figure}

Recent work \citep{jeong2023winclip,zhou2023anomalyclip} leverages the zero-shot generalization ability of foundation models \citep{radford2021learning} to address anomaly segmentation across a varied number of objects and environments.
A modern line of work in computer vision deals with test-time robustness of discriminative models against natural distribution shifts \citep{hendrycks2021natural,chen2023advdiffuser}, where it is found that, generally, even large-scale models are fragile to well-chosen, per-sample perturbations.
This has been recently demonstrated for CLIP-based models both for the case of noise-like perturbations \citep{schlarmann2024robust} and for natural perturbations (such as camera pose changes) \citep{tong2024eyes}.

In this work, we investigate the robustness of downstream anomaly segmentation using CLIP when faced with three perturbations (rotations, hue shift, saturation shift) that simulate naturally occurring distribution shifts.
Importantly, we bound these shifts in a controlled way such that anomalies remain clearly present when inspected by humans -- the question we investigate is \textit{if CLIP-based anomaly segmentation is invariant to these shifts}?

Our results demonstrate that average segmentation performance is reduced across most objects across CLIP backbones, with drops of up to $40\%$ in the per-sample worst-case setting.
This performance drop points to the need of conducting lower bound performance evaluations of foundation models used in zero-shot anomaly segmentation, and is in line with a recent line of work that investigates the resilience of foundation models to natural distribution shifts and adversarial examples \citep{kandpal2023backdoor,shi2023effective}.

\subsection{Related Work}
A large number of recent works have introduced zero-shot anomaly classification and segmentation algorithms that use foundation models \citep{huang2022registration,gu2024anomalygpt,cao2023segment, cao2023towards,wyatt2022anoddpm}.
Among these, we focus on WinCLIP \citep{jeong2023winclip} because of its good performance, fast inference speed, and open-source implementation in the \texttt{anomalib} library \citep{akcay2022anomalib}.
WinCLIP for anomaly segmentation has two components: a template-based prompting strategy, and a multi-scale pooling strategy to aggregate information across multiple spatial scales.
Other recent methods have used the properties of diffusion processes for anomaly detection \citep{livernoche2023diffusion}, pre-trained diffusion models \citep{wyatt2022anoddpm}, and vision-language models \citep{gu2024anomalygpt}.

Recent work \citep{shi2023effective,tu2024closer} investigates the robustness of CLIP and how previously reported results likely over-state the observed effective robustness due to evaluating on a static test set from one distribution. The work in \citet{idrissi2022imagenet} characterizes subsets of the ImageNet dataset and identifies color shifts as a common subset and potential augmentation.

Our proposed work is similar in spirit to both previous idea, given that we propose to evaluate anomaly segmentation on worst-case augmented samples from the original test set, and our augmentations include color shifts \citep{hosseini2018semantic,bhattad2020unrestricted}.
Another line of work \citep{lewis2023does,tong2024eyes,udandarao2023visual} questions whether CLIP suffers from inaccurate physical grounding, showing that semantic properties of images are not understood by vision-language models such as LLaVA \citep{liu2023visual} or GPT-4V \citep{openai2024gpt4}.

\section{Methods}
\label{sec:methods}

To investigate the performance of anomaly segmentation under worst-case distribution shifts, we use three semantic preserving augmentations: rotation, hue shift, and saturation shift. Their effects on the same sample are shown in Figure~\ref{fig:main_aug_examples}, where it can be seen that the anomaly remains detectable across the entire augmentation range.

The query sample $x$ is clockwise rotated using an angle $\theta$, followed by an RGB to HSV conversion.
Additive shifts $\delta_h$ and $\delta_s$ are independently applied to the hue and saturation channels \citep{hosseini2018semantic}, respectively, after which the sample is converted back to its RGB representation, yielding the augmented sample $x_\textrm{aug}$.
The pre-processing is summarized as:
\begin{equation}
    x_\textrm{aug} = f(x; \theta, \delta_h, \delta_s).
\end{equation}

\noindent The modified sample $x_\textrm{aug}$ is input to zero-shot anomaly detection using WinCLIP \citep{jeong2023winclip}, yielding the estimated soft (continuous) anomaly maps $\tilde{y}_\textrm{aug}$ as:
\begin{equation}
    \tilde{y}_\textrm{aug} = w(x_\textrm{aug}).    
\end{equation}

\noindent Finally, the output sensitivity maps are additionally rotated counter-clockwise using the angle $\theta$ (hue and saturation shifts do not spatially change the sample) to yield:
\begin{equation}
    \tilde{y} = f(\tilde{y}_\textrm{aug}; -\theta, 0, 0),
\end{equation}

\noindent where the choice of $\delta_h = \delta_s = 0$ implies there is no HSV space conversion performed.
This allows for fair end-to-end comparison between $\tilde{y}$ and the ground-truth segmentation maps $y$, without requiring any intervention on $y$.

The entire pre-processing pipeline is differentiable, thus the parameters $\theta, \delta_h, \delta_s$ can be optimized using gradient-based methods. 
However, doing so on a per-sample basis requires defining a differentiable loss function that measures anomaly segmentation loss. 
We introduce the following simplified version of the Dice loss \citep{Sudre_2017} (using the $\ell_1$-norm) from medical image segmentation:
\begin{equation}
    l(\tilde{y}, y) = \frac{\sum_i \tilde{y}_i (1 - y_i)}{\sum_i (1 - y_i) + \epsilon} - \frac{\sum_i \tilde{y}_i y_i}{\sum_i y_i + \epsilon},
\label{eq:our_loss}
\end{equation}

\noindent where $\epsilon = 1\mathrm{e}\textrm{-}8$ is a small constant added for numerical stability.

Note that this is a valid segmentation loss: the first term encourages low values for $\tilde{y}_i$, where $i$ satisfies $y_i = 0$ (i.e., true negative pixels). 
Similarly, the second term encourages higher values of $\tilde{y}_i$ for the true positive pixels.
Unlike the Dice loss, our proposed loss does not contain the estimated output $\tilde{y}$ in the denominator, which improves numerical stability when back-propagating gradients to the inputs.

Unlike supervised learning (where the goal is loss minimization), the goal of our work is to \textbf{\textit{maximize}} this loss by optimizing the three parameters $\theta, \delta_h, \delta_s$ as:
\begin{equation}
\begin{aligned}
    \theta^*, \delta_h^*, \delta_s^* \ = \ & \argmax_{\theta, \delta_h, \delta_s} \ l\left(\tilde{y}(\theta, \delta_h, \delta_s), y\right), \\
    \textrm{s.t.} & \ \theta \in \left[-90^\circ, 90^\circ\right], \delta_s \in \left[-0.5, 0.5\right].
\end{aligned}
\label{eq:adv_optimization}
\end{equation}

\begin{figure}[t]
    \centering
    \includegraphics[width=1\linewidth]{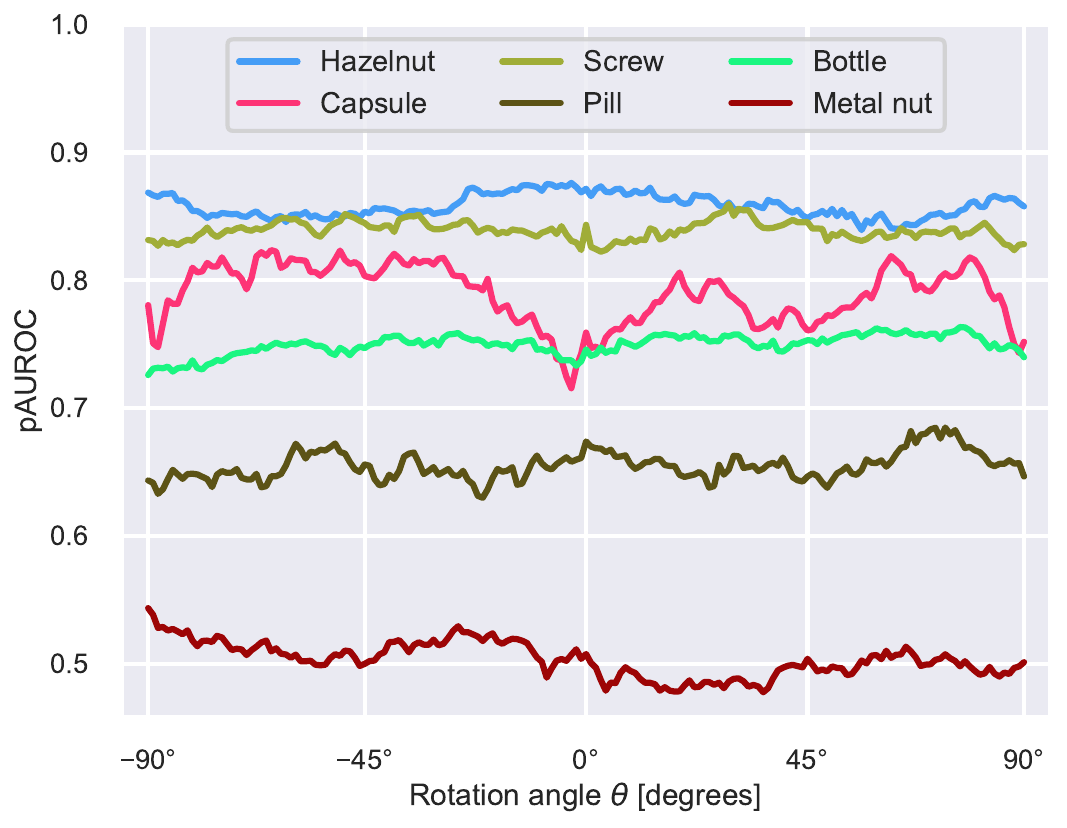}
    \caption{Zero-shot anomaly segmentation performance when the same rotation angle $\theta$ is applied to the MVTec test set for rotation-invariant objects.}
    \label{fig:rot_grid_mvtec}
\end{figure}

\begin{figure}[t]
    \centering
    \includegraphics[width=1\linewidth]{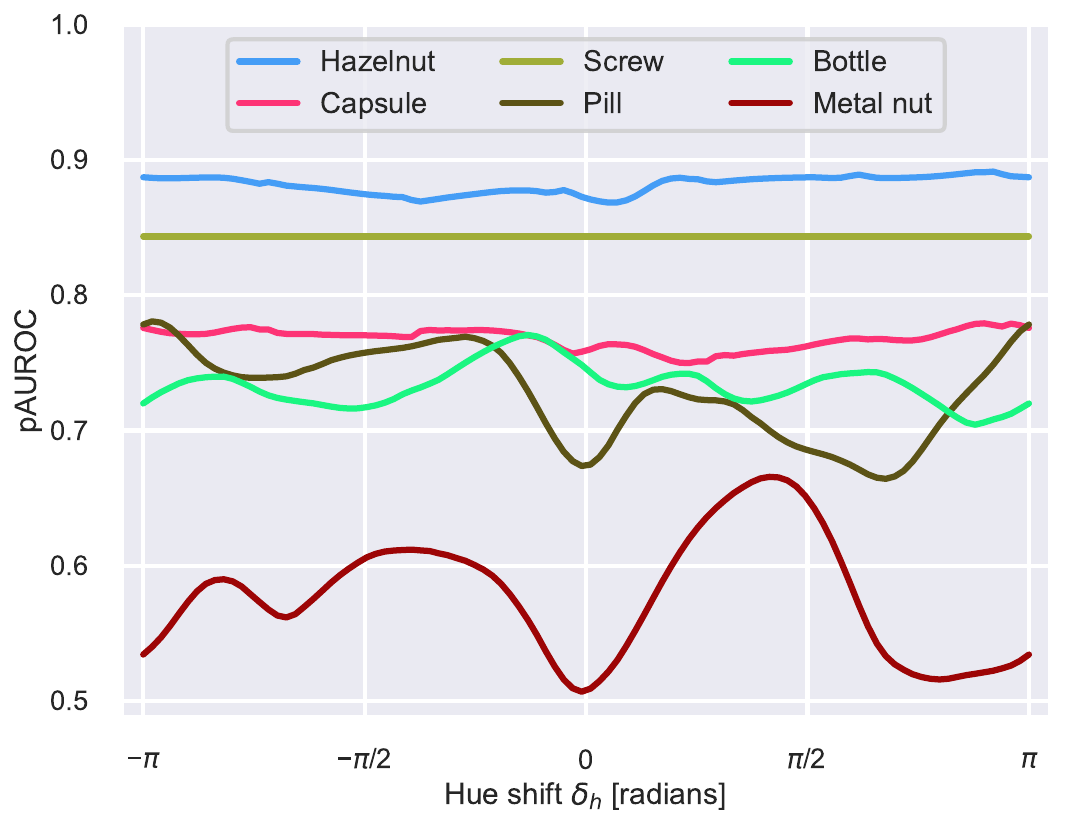}
    \caption{Zero-shot anomaly segmentation performance when the same additive (modulo $2\pi$) hue shift $\delta_h$ is applied to the MVTec test set for rotation-invariant objects.}
    \label{fig:hue_grid_mvtec_rot}
\end{figure}

\noindent To obtain the tightest lower performance bounds, this optimization is done separately for each sample in the test set. We used the Adam optimizer \cite{kingma2014adam} with at most $200$ optimization steps per sample to approximate the solution to \eqref{eq:adv_optimization}. Details are provided in the supplementary material.

The above is an approximation of the worst-case setting encountered in practice across an extended period of time: every individual test sample is manipulated in its own worst way (by accident or ill-intended) with respect to the downstream task, while still ensuring that the manipulation is realistic and preserves the semantic information of the sample (e.g., using a bounded saturation shift). Samples are aggregated and final performance is measured using three metrics, following the protocol described in \citet{jeong2023winclip}: pixel-level area under the ROC curve (pAUROC), area under the per-region overlap curve (AUPRO), and the $F_1$-max score.

\section{Experimental Results and Discussion}

\begin{figure*}[t]
    \centering
    \includegraphics[width=1\linewidth]{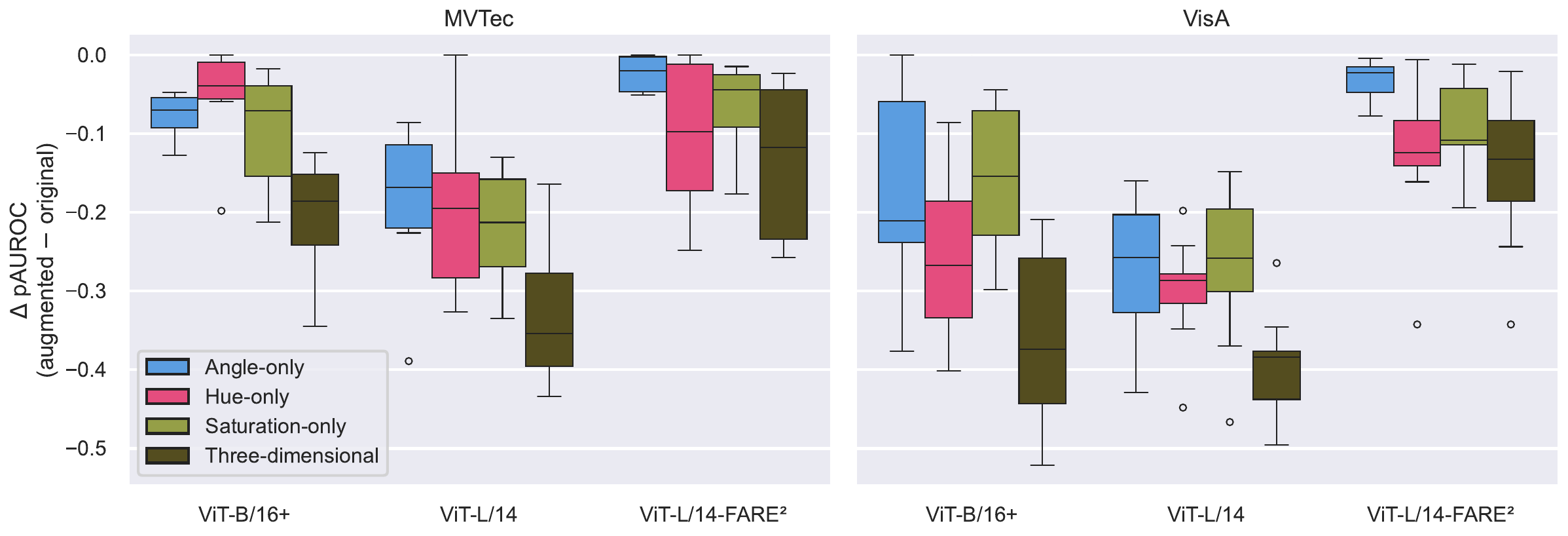}
    \caption{Zero-shot anomaly segmentation performance for two datasets (MVTec and VisA, left and right, respectively) using three CLIP backbones (ViT-B/16+, ViT-L/14, and adversarially fine-tuned ViT-L/14-FARE$^2$ for WinCLIP. The three test-time, worst-case semantic perturbations (angle, saturation, hue) are considered either separately, or simultaneously (3D). The bars show the difference between the original test sets and the considered lower bounds.}
    \label{fig:aggregated_results}
\end{figure*}

\begin{table*}
\centering
\begin{tabularx}{\linewidth}{Xccccccccc}
\toprule
\multicolumn{2}{c}{Anomaly Segmentation} & \multicolumn{4}{c}{MVTec-AD} & \multicolumn{4}{c}{VisA} \\
\cmidrule(r){3-6} \cmidrule(l){7-10}
Eval. data & Backbone & pAUROC & AUPRO & $F_1$-max & Loss & pAUROC & AUPRO & $F_1$-max & Loss \\
\midrule
\multirow{3}{*}{Original}
      & ViT-B/16+ & 80.1 & 64.2 & 24.1 & 0.075 & 76.4 & 55.4 & 10.9 & 0.165 \\
      & ViT-L/14 & 59.1 & 21.9 & 9.9 & 0.108 & 59.3 & 26.3 & 2.89 & 0.199 \\
      & ViT-L/14-$\text{FARE}^2$ & 37.7 & 10.6 & 8.3 & 0.108 & 30.6 & 4.81 & 2.10 & 0.203 \\
\midrule
      Per-sample & ViT-B/16+ & 59.8 & 23.6 & 12.8 & 0.085 & 40.3 & 16.0 & 2.28 & 0.180 \\
      3D lower & ViT-L/14 & 26.4 & 4.25 & 7.68 & 0.115 & 19.9 & 3.59 & 1.37 & 0.210 \\
      bound & ViT-L/14-$\text{FARE}^2$ & 24.2 & 3.54 & 7.80 & 0.111 & 16.5 & 2.12 & 1.37 & 0.208 \\
\bottomrule
\end{tabularx}
\caption{Anomaly segmentation performance under the original test data and the per-sample worst-case lower bound.}
\label{tab:average_results}
\end{table*}

\subsection{Uniform lower performance bounds}

We first evaluate the performance of zero-shot anomaly segmentation when only one of the three augmentations is applied \textit{using the same value} to the entire test set. 
This is a loose lower bound, but we find that even in this case, conclusions are non-trivial.

Figure~\ref{fig:rot_grid_mvtec} shows zero-shot segmentation pAUROC for six objects in the MVTec dataset when the same rotation angle $\theta$ is applied to all test samples using the protocol in Section~\ref{sec:methods}. 
Only objects that are placed on matte backgrounds were chosen for evaluation, to ensure that rotation does not produce artificial anomalies in the corners of $x_\textrm{aug}$.
Figure~\ref{fig:rot_grid_mvtec} reveals that rotating samples has a non-trivial and non-smooth effect on downstream task performance: small rotations at arbitrary angles can cause a performance difference of up to $3\%$ in pAUROC to appear in most objects.
Performance can vary by up to $10\%$ between the lower and upper bounds of a certain object (e.g., for capsule).
Similar results for the VisA dataset are in the supplementary material.

Figure~\ref{fig:hue_grid_mvtec_rot} shows zero-shot segmentation pAUROC the same six objects in Figure~\ref{fig:rot_grid_mvtec} when the same hue shift $\delta_h$ is applied to the test samples.
In this case, variations are smoother, but effects are still non-trivial: certain objects are less sensitive (e.g., all screw images are gray-scale, and completely unaffected by hue), while others show a significant performance variation (e.g., pill and metal nut) of up to $15\%$.
Given that hue shifts should not affect physical defects in objects, this indicates that the backbone used here (ViT-B/16+) is not sufficiently robust to these shifts, even when the shifts are not sample-specific.

\subsection{Per-sample lower performance bounds}
We report per-sample lower performance bounds where the worst-case transformation is independently obtained for each test sample, followed by aggregating all augmented samples and reporting performance.

As expected from Figure~\ref{fig:rot_grid_mvtec}, the distribution of worst-case angles across individual samples is approximately uniform: given that only rotation-invariant objects were evaluated, there is no reason for specific angles to be systematically worst-case for all samples. 
The distribution is shown in the supplementary material.

Figure~\ref{fig:aggregated_results} shows the distribution of performance drops across objects for three CLIP backbones tested on both the augmented MVTec and VisA datasets.
Four per-sample worst-case augmentations were used: three one-dimensional ones -- in angle $\theta$, hue shift $\delta_h$, saturation shift $\delta_s$ -- and the worst-case lower performance bound obtained by jointly optimizing all three variables for each sample and aggregating results.
In general, we notice that the lower bound is looser (performance suffers a larger drop) for the VisA dataset compared to MVTec in the case of non-adversarially fine-tuned backbones. 
When considering that VisA is a more difficult test set that MVTec (based on the lower original performance in Table~\ref{tab:average_results}), the even higher worst-case performance drop indicates that a significant robustness gap exists in harder segmentation problems.

In terms of specific one-dimensional perturbations, hue shifts affect VisA the most on average (across all backbones), while the results for MVTec are less conclusive, and more backbone-dependent.
The adversarially fine-tuned FARE$^2$ backbone \citep{schlarmann2024robust} consistently shows lower performance drops, but the nominal performance also suffers a significant drop, indicating a sharp trade-off between robustness to distribution shifts and absolute performance.

Table~\ref{tab:average_results} shows results for anomaly segmentation metrics, evaluated on both datasets, using three models.
We compare the original (un-augmented) test set performance with that of the worst-case per-sample lower bound.
Given the imbalanced nature of anomaly segmentation (fewer positive pixels than negatives), the AUPRO scores suffers a performance of up to $40\%$ using the ViT-B/16+ architecture.
Similarly, we reproduce the already-low F1-max score on the original test sets, and this value drop to as low as $2\%$ for the VisA dataset using the ViT-B/16+ architecture.
Table~\ref{tab:average_results} shows the numerical values of the optimization objective defined in \eqref{eq:our_loss} used to find worst-case augmentations: it can be seen that the values are inversely correlated with the performance metrics, and that \eqref{eq:our_loss} is a valid segmentation loss.

\section{Conclusion}
Our work demonstrates the generalization challenges of zero-shot anomaly segmentation algorithms when faced with relatively simple test-time augmentations (rotation, hue, saturation). By formulating an optimization problem that determines the per-sample worst-case augmentations, we show performance drops for CLIP-based anomaly segmentation of up to $40\%$ in pAUROC compared to the original test set.

Standard practice for CLIP pre-trained models is to not use heavy augmentation pipelines \citep{radford2021learning} -- some samples have text label information about their color, which would make naive color augmentations not compatible with the learning objective. Future research could consider how to generalize semantic augmentations to multi-modal data and include this in the training of foundation models. Finally, all our work targeted zero-shot anomaly segmentation. Few-shot anomaly segmentation is common in practice and significant improvements could be achieved if test-time augmentations were optimized to \textit{maximize} performance for each object type instead of minimizing it, as we did in this work.


{
    \small
    \bibliographystyle{ieeenat_fullname}
    \bibliography{main}
}


\clearpage
\setcounter{page}{1}
\setcounter{section}{0}

\maketitlesupplementary
\renewcommand\thesection{\Alph{section}}

\section{Implementation Details}

\subsection{Augmentations}
\label{subsec:augs_appendix}
To implement the rotation augmentation, we used the three-pass convolutional approach from \citet{unser1995convolution}. Compared to bilinear interpolation, this method preserves the high-frequency content of the samples and ensures that minimal distortions (at least when evaluated by humans) are introduced via the rotation. To extrapolate pixels in the corners of the rotated sample, we used the \texttt{replicate} padding mode: for each extrapolated pixel, the value of the nearest pixel from the rotated sample itself is copied. We used the \texttt{torch-rotation} library \citep{torch_rotation} to implement the differentiable three-pass rotation.

Hue and saturation shifts were implemented using the RGB-to-HSV conversion function from the \texttt{kornia} library \citep{eriba2019kornia}, with the shifts themselves being otherwise straightforward to implement. 

Given that saturation is clipped to $[0, 1]$ after the shift is applied, we used a straight-through estimator (STE) \citep{bengio2013estimating} to back-propagate through this operation even outside its valid range. Similarly, we used the STE estimator to clip and back-propagate the values of the rotation angle $\theta$ and saturation shift $\delta_s$ whenever they exceed their permitted ranges.

\subsection{Solving the optimization problem}
We used the Adam optimizer \citep{kingma2014adam} with learning rates of $5$ for $\theta$ and $0.1$ for $\delta_h$ and $\delta_s$, respectively. The learning rates were chosen based on the $\ell_\infty$-norm of the gradient with respect to each variable at initialization, and were not extensively fine-tuned or searched for.

We found that using random restarts and for the optimization problem is crucial for successfully decreasing the per-sample lower performance bound. In particular, the loss surface is highly non-convex for the angle $\theta$, and optimization often gets trapped in local minima. Given the relatively low budget of $200$ steps, we used $5$ restarts: the first optimization run starts with the default values $\theta = \delta_h = \delta_s = 0$, while the subsequent four runs restart these values using uniform sampling from the valid interval of each variable.

Additionally, we also monitor the per-sample segmentation pAUROC for samples that have at least one anomalous labeled pixel in $y$ and select the specific perturbations $(\theta^*, \delta_h^*, \delta_s^*)$ that minimize this value. Note that this is still sub-optimal when it comes to minimizing the pAUROC on pixels from all samples of a specific object, which is generally intractable. Finally, for samples that have $y_i = 0$ for all $i$ (no anomalous pixels), we monitor the loss function $L$ in \eqref{eq:our_loss} and select the specific perturbations $(\theta^*, \delta_h^*, \delta_s^*)$ that maximize this value.

\subsection{Selecting rotation-invariant objects}
To select rotation-invariant objects, we used the per-pixel average mean squared error between the extrapolated corner pixels using our rotation method and the original corner pixels as:
\begin{equation}
    e(x) = \frac{\sum_i m_i \cdot (x_i - f(x_i; \theta=45^\circ))^2}{\sum_i m_i},
\end{equation}
\noindent where $m_i$ is a binary mask that indicates whether a specific pixel was extrapolated or not and is obtained by rotating a monochrome image with maximum pixel intensity. We evaluate this error using the specific angle $\theta = 45^\circ$, with the reasoning that it is where the most extrapolation occurs, and rotation-invariant objects should not be cut-off by this rotation. If a cut-off would occur, the object texture would be reflected in the extrapolated corner, thus producing a high error compared to the original corner region.

The above is followed by a manual thresholding of the error, as well as final careful visual inspection of the down-selected classes. Tables~\ref{tab:mvtec_corner_results} and~\ref{tab:visa_corner_results} show the selection results for objects in the MVTec and VisA datasets, respectively.

\begin{table}[ht]
\centering
\begin{tabularx}{0.9\linewidth}{XXX}
\hline
\textbf{Object name} & \textbf{Corner MSE} & \textbf{Used} \\
\hline
Metal nut & 0.000255 & True \\
Pill & 0.000304 & True \\
Hazelnut & 0.000391 & True \\
Bottle & 0.002179 & True \\
Screw & 0.005964 & True \\
Leather & 0.013949 & False \\
Wood & 0.021015 & False \\
Capsule & 0.02621 & True \\
Cable & 0.027923 & False \\
Transistor & 0.077574 & False \\
Tile & 0.09312 & False \\
Carpet & 0.104486 & False \\
Grid & 0.126188 & False \\
Toothbrush & 0.17937 & False \\
Zipper & 0.92926 & False \\
\hline
\end{tabularx}
\caption{Average corner extrapolation MSE for objects in the MVTec dataset. \textit{Leather} and \textit{wood} were excluded from evaluation after manual inspection, given that the textures span the entire sample and simple extrapolation would lead to qualitative artifacts in the samples that could be mistaken for anomalies.}
\label{tab:mvtec_corner_results}
\end{table}

\begin{table}[ht]
\centering
\begin{tabularx}{0.9\linewidth}{XXX}
\hline
\textbf{Object name} & \textbf{Corner MSE} & \textbf{Used} \\
\hline
Pipe fryum & 0.000668 & True \\
Pcb3 & 0.001449 & True \\
Macaroni1 & 0.004837 & True \\
Pcb4 & 0.005264 & True \\
Fryum & 0.00995 & True \\
Pcb2 & 0.010521 & True \\
Pcb1 & 0.013953 & True \\
Chewing gum & 0.024117 & True \\
Macaroni2 & 0.035739 & True \\
Cashew & 0.049281 & True \\
Candle & 0.051116 & False \\
Capsules & 0.082838 & False \\
\hline
\end{tabularx}
\caption{Average corner extrapolation MSE for objects in the VisA dataset.}
\label{tab:visa_corner_results}
\end{table}

\subsection{Optimization outcomes}
\label{sec:supp_opt_outcomes}

\begin{figure}[t]
    \centering
    \includegraphics[width=1\linewidth]{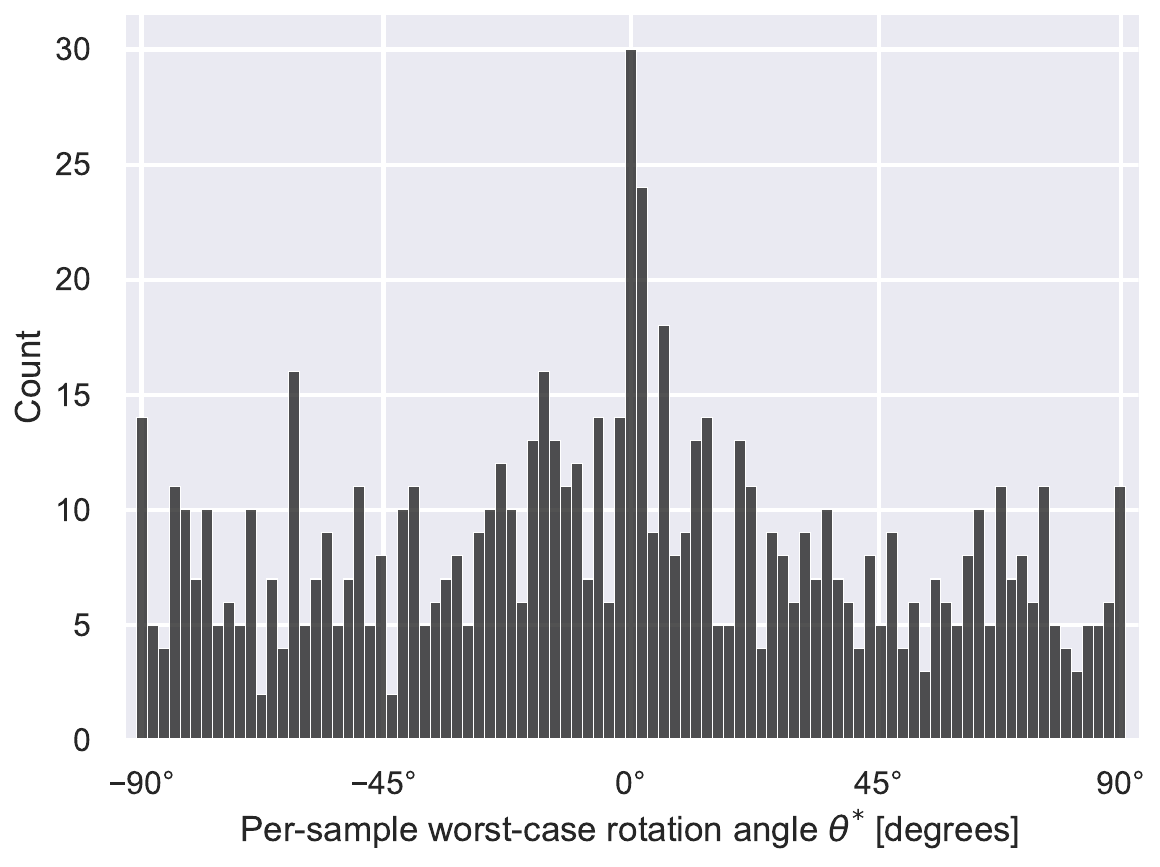}
    \caption{Empirical distribution of per-sample worst-case rotation angles for all rotationally invariant object in MVTec. The slight peak around the origin is caused by the optimization being sub-optimal due to the de-rotation of $\tilde{y}_\textrm{aug}$ using zero value extrapolation in the corners.}
    \label{fig:worst_case_angle_all}
\end{figure}

Figure~\ref{fig:worst_case_angle_all} shows the histogram of per-sample worst-case rotation angles $\theta^*$ when optimizing only for rotation.
The higher mode around the origin is due to the intentional artifacts introduced when de-rotating the estimated anomaly map $\tilde{y}_\textrm{aug}$: for difficult samples, this may cause a slight performance boost at any rotation angle.
This causes the optimization in ~\eqref{eq:adv_optimization} to return a near-zero angle as the worst-case one.

\section{Additional uniform lower bound results}
\begin{figure}[t]
    \centering
    \includegraphics[width=1\linewidth]{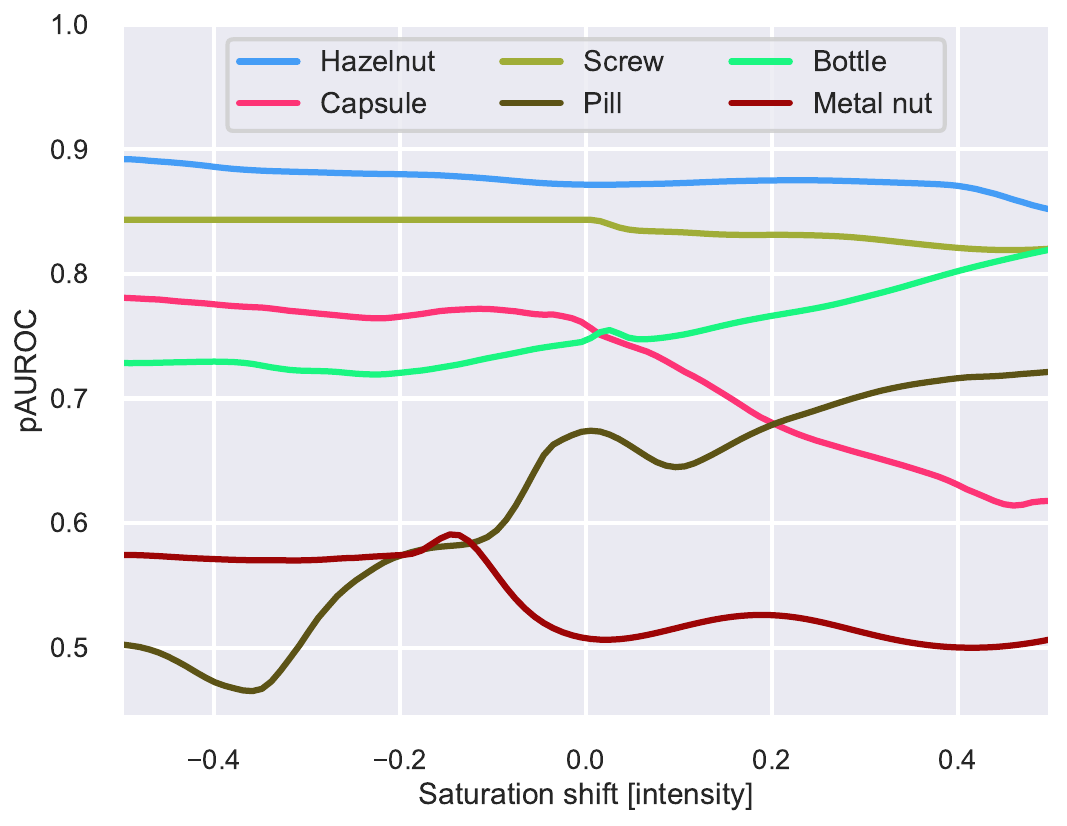}
    \caption{Zero-shot anomaly segmentation performance when the same saturation shift $\delta_s$ is applied to the MVTec test set for rotation-invariant objects.}
    \label{fig:sat_grid_mvtec}
\end{figure}

Figure~\ref{fig:sat_grid_mvtec} shows the results for applying the same saturation shift $\delta_s$ to all samples in the MVTec test set. Similar to the result for the hue shift $\delta_h$ in Figure~\ref{fig:hue_grid_mvtec_rot}, there are significant performance variations for most objects except metal nut, which already contains gray-scale samples that are less sensitive to color shifts.

\begin{figure}[t]
    \centering
    \includegraphics[width=1\linewidth]{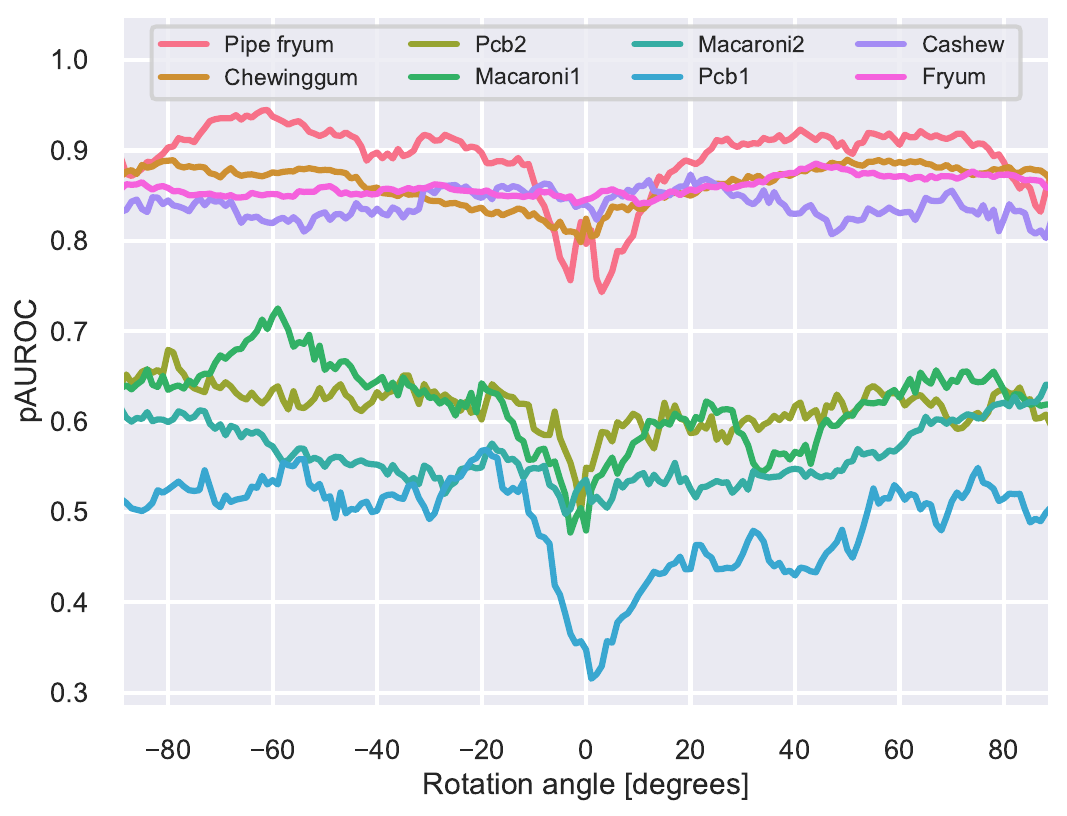}
    \caption{Zero-shot anomaly segmentation performance when the same rotation angle $\theta$ is applied to the VisA test set for rotation-invariant objects.}
    \label{fig:rot_grid_visa}
\end{figure}

\begin{figure}[t]
    \centering
    \includegraphics[width=1\linewidth]{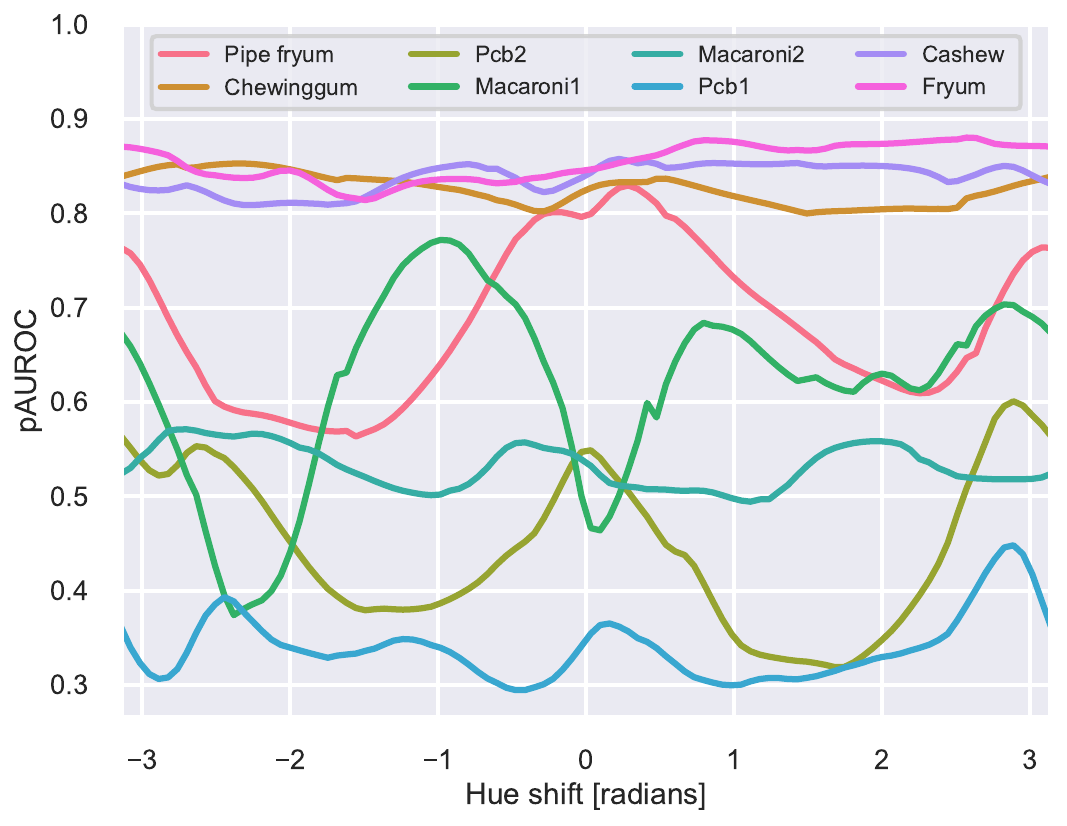}
     \caption{Zero-shot anomaly segmentation performance when the same hue shift $\delta_h$ is applied to the VisA test set for rotation-invariant objects.}
    \label{fig:hue_grid_visa}
\end{figure}

\begin{figure}[t]
    \centering
    \includegraphics[width=1\linewidth]{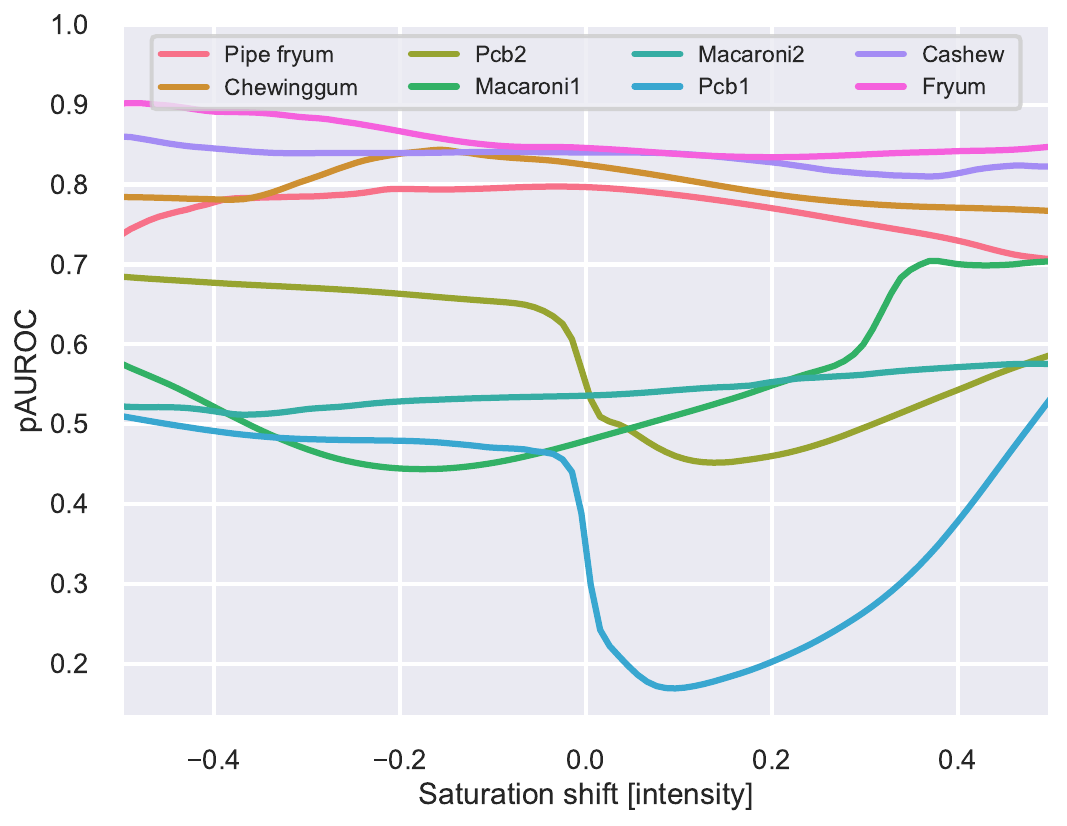}
    \caption{Zero-shot anomaly segmentation performance when the same saturation shift $\delta_s$ is applied to the VisA test set for rotation-invariant objects.}
    \label{fig:sat_grid_visa}
\end{figure}

Figures~\ref{fig:rot_grid_visa}, \ref{fig:hue_grid_visa}, and~\ref{fig:sat_grid_visa} show the uniform lower bounds obtained from applying the same rotation, hue shift, or saturation shift to all test samples in the VisA dataset, respectively. While the same conclusions related to how smooth performance varies with each augmentation type (noisy for rotations, smooth for color shifts), the dynamic ranges of the per-object performance are much larger for VisA, which is generally a more difficult dataset. For certain objects, such as \texttt{Pcb1} or \texttt{Macaroni1}, performance can vary by up to $0.4$ points in pAUROC when rotations or hue shifts are applied, respectively.

\section{Additional worst-case lower bound results}
\begin{figure*}[t]
    \centering
    \includegraphics[width=1\linewidth]{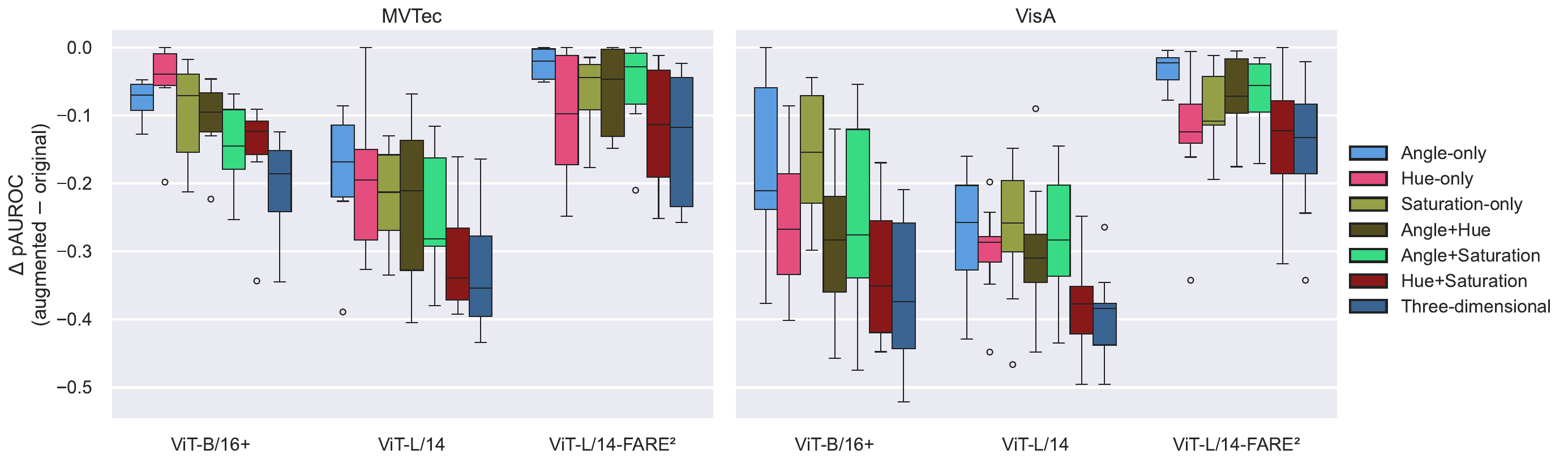}
    \caption{Zero-shot anomaly segmentation performance for two datasets (MVTec and VisA, left and right, respectively) using three CLIP backbones (ViT-B/16+, ViT-L/14, and adversarially fine-tuned ViT-L/14-FARE$^2$ for WinCLIP. Each possible combination of the three test-time, worst-case semantic perturbations (angle, saturation, hue) is considered, to yield a set of eight total test-time augmentation strategies. The bars show the difference between the original test sets and the considered lower bounds. This figure complements Figure~\ref{fig:aggregated_results} in the main body.}
    \label{fig:aggregated_results_supp}
\end{figure*}

Figure~\ref{fig:aggregated_results_supp} shows the results for worst-case lower bounds obtained by jointly optimizing any combination of considered augmentations. In general, we note that augmentations that only involve colours (hue and saturation) are more effective (leads to a higher performance drop) than the ones involving the rotation augmentation. We attribute this to the non-smooth optimization surface for the rotation angle $\theta$, which holds even under our considered segmentation loss.

\end{document}

%% file: preamble.tex
\usepackage[dvipsnames]{xcolor}
